\documentclass[10pt,twocolumn,letterpaper]{article}

\usepackage{cvpr}              

\usepackage{tcolorbox}

\definecolor{cvprblue}{rgb}{0.21,0.49,0.74}
\usepackage[pagebackref,breaklinks,colorlinks,allcolors=cvprblue]{hyperref}
\usepackage{multirow}

\usepackage{pgfplots}
\usepackage{times}
\usepackage{pgfplotstable}
\pgfplotsset{compat=1.18}

\usepackage{xcolor}         
\usepackage{colortbl}
\def\paperID{23994} 
\def\confName{CVPR}
\def\confYear{2026}

\title{MSRL: Scaling Generative Multimodal Reward Modeling \\ via Multi-Stage Reinforcement Learning}

\author{
\textbf{Chenglong Wang\textsuperscript{1}\thanks{This work was done while interning at ByteDance.},
Yifu Huo\textsuperscript{1}$^*$,
Yang Gan\textsuperscript{1},
Qiaozhi He\textsuperscript{2}}, \\
\textbf{Qi Meng\textsuperscript{1},
Bei Li\textsuperscript{1},
Yan Wang\textsuperscript{2},
Junfu Liu\textsuperscript{2},
Tianhua Zhou\textsuperscript{2},
Jingbo Zhu\textsuperscript{1}
and Tong Xiao\textsuperscript{1}\thanks{Corresponding author.}}\\
\textsuperscript{1}Northeastern University, Shenyang, China \\
\textsuperscript{2}ByteDance \\
\tt{clwang1119@gmail.com \quad xiaotong@mail.neu.edu.cn}
}

\begin{document}
\maketitle

\begin{abstract}
Recent advances in multimodal reward modeling have been largely driven by a paradigm shift from discriminative to generative approaches.
Building on this progress, recent studies have further employed reinforcement learning from verifiable rewards (RLVR) to enhance multimodal reward models (MRMs).
Despite their success, RLVR-based training typically relies on labeled multimodal preference data, which are costly and labor-intensive to obtain, making it difficult to scale MRM training.
To overcome this limitation, we propose a \textbf{\underline{M}}ulti-\textbf{\underline{S}}tage \textbf{\underline{R}}einforcement \textbf{\underline{L}}earning (MSRL) approach, which can achieve scalable RL for MRMs with limited multimodal data. MSRL replaces the conventional RLVR-based training paradigm by first learning a generalizable reward reasoning capability from large-scale textual preference data, and then progressively transferring this capability to multimodal tasks through caption-based and fully multimodal reinforcement-learning stages.
Furthermore, we introduce a cross-modal knowledge distillation approach to improve preference generalization within MSRL.
Extensive experiments demonstrate that MSRL effectively scales the RLVR-based training of generative MRMs and substantially improves their performance across both visual understanding and visual generation tasks (\textit{e.g.,} 66.6\%$\rightarrow$75.9\% on VL-RewardBench and 70.2\%$\rightarrow$75.7\% on GenAI-Bench), without requiring additional multimodal preference annotations. Our code is available at: https://github.com/wangclnlp/MSRL.

\end{abstract}

\begin{figure}[!t]
    \centering
    \begin{subfigure}{\linewidth}
        \centering
        \includegraphics[width=\linewidth]{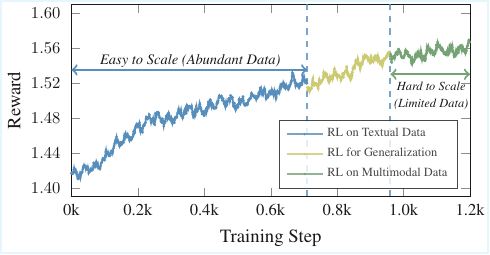}
        
        \hspace{0.98cm}
        \begin{minipage}{0.8\linewidth}
        \vspace{-4mm}
        \caption{Learning Curves across Modalities}
        \end{minipage}
    \end{subfigure}
    \begin{subfigure}{\linewidth}
        \centering
        \vspace{0.2cm}
        \includegraphics[width=\linewidth]{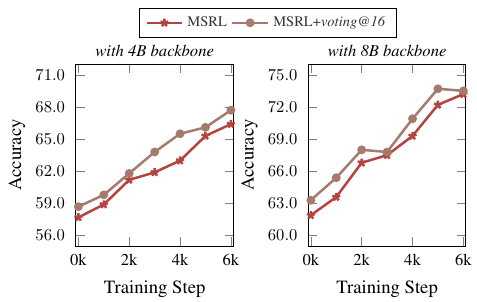}
        \caption{Accuracies (\%) of MRMs across Training Steps on Textual Data}
        \label{fig:intro_b}
    \end{subfigure}
    \caption{
    Illustration of our multi-stage RL approach. Subfigure (a) shows that abundant textual preference data can facilitate scalable RL. Subfigure (b) demonstrates that we can effectively scale multimodal generative reward models through multi-stage RL.
     }
    \vspace{-0.3cm}
    \label{fig:scaling_in_intro}
\end{figure}

\section{Introduction}
One of the central drivers behind the rapid progress of multimodal large language models (MLLMs) in recent years has been the application of \textit{human preference alignment}, which enables MLLMs to generate outputs that are more helpful, safe, and factually grounded \cite{ouyang2022training,wang2024lift,wang2024hybrid,wang2024esrl,zhou2024prior,liu2025videodpo,pu2025judge,xiong2025llava,huo2025heal}. A commonly used approach is to learn a multimodal reward model (MRM) to encode human preferences, and then fine-tune the MLLM to maximize the reward signal it provides \cite{yasunaga2025multimodal,wang2025skywork,zhang2025structvrm,wang2025probing}. More recently, the role of MRMs has extended beyond training to inference, where they are used to re-rank candidate outputs. This approach has emerged as a strategy in studies on inference time scaling laws \cite{xiao2025foundations,wang2025visualprm,zhang2025survey,snell2025scaling,chang2025step}.

The standard paradigm for developing MRMs has recently shifted from discriminative to generative reward modeling, where reward prediction is formulated as a chain-of-thought (CoT) reasoning and text generation process \cite{zhang2024generative,mahan2024generative,wang2025rovrm,xiong2025llava,wang2025gram}.  
Building on this progress, a more sophisticated line of research integrates RL into MRM, exemplified by reinforcement learning with verifiable rewards (RLVR) \cite{zhang2025r1,wang2025unified}. 
However, this line of progress faces a critical bottleneck: the profound scarcity of high-quality, human-annotated \textit{multimodal} preference data. The data-hungry nature of RLVR, which relies on explicit preference labels for verifiable rewards, makes it difficult to scale \cite{liu2025prorl, guo2025deepseek}. This so-called ``data bottleneck'' is not a minor inconvenience, it is a fundamental barrier. It prevents the field from leveraging a key insight observed in text-only LLMs: that scaling RL can significantly enhance the underlying reasoning capabilities of the models themselves \cite{liu2025prorl, hu2025brorl, guo2025deepseek}.

To date, existing solutions have proven insufficient for overcoming this bottleneck. The most direct approach, annotating massive new multimodal preference datasets, is prohibitively expensive. Alternative strategies, such as constructing reward signals from confidence estimation \cite{prabhudesai2025maximizing} or self-verification \cite{liu2025trust, wang2025vl}, often suffer from error propagation and quickly hit performance plateaus. We are thus left with a crucial question: \textit{how can we effectively scale RL for MRMs without an unscalable annotation budget?}

In this work, we challenge the underlying hypothesis that this bottleneck must be solved with \textit{more multimodal data}. We propose a different perspective: the core \textit{reasoning} abilities for preference modeling can be learned from abundant, text-only data and effectively transferred across modalities, building on recent findings that preferences can generalize \cite{wang2025rovrm}. This insight forms the basis of our approach: \textbf{\underline{M}}ulti-\textbf{\underline{S}}tage \textbf{\underline{R}}einforcement \textbf{\underline{L}}earning (MSRL), a scalable curriculum that leverages vast textual preference data to mitigate the multimodal data bottleneck. MSRL operates in multi-stages: it first learns a generalizable reward reasoning ability from large-scale textual preference data, and then progressively transfers and specializes this ability to multimodal tasks, first via caption-based RL and finally with fully multimodal RL using the limited available preference data. To further enhance this cross-modal generalization, we introduce a Cross-Modal Knowledge Distillation (CMKD) technique, which scalably aligns representations without requiring additional unlabeled preference data.

Our MSRL is effective across diverse multimodal tasks.  In our experiments, we apply MSRL to train a unified generative MRM \cite{wang2025unified}, which can be used to predict preferences for both visual understanding and visual generation tasks. Such a model is particularly challenging to build due to the scarcity of high-quality mixed-modality preference data. Our experimental results demonstrate that integrating RL on textual data can effectively scale multimodal reward modeling and substantially improve the accuracy of MRMs (\textit{e.g.,} 66.6\%$\rightarrow$75.9\% on VL-RewardBench and 76.2\%$\rightarrow$80.5\% on Multimodal RewardBench). Beyond improvements on individual tasks, MSRL exhibits consistent scaling behavior across different model sizes (see Figures~\ref{fig:scaling_in_intro} and ~\ref{fig:scaling_res_with_diff_model_sizes}): from 1B to 14B backbones, larger models benefit even more substantially from textual RL, further highlighting the robustness and practicality of our approach for scaling generative multimodal reward modeling.

\section{Related Work}
\subsection{Multimodal Large Language Models}
Inspired by the success of large language models (LLMs) such as GPTs \cite{ouyang2022training} and LLaMA \cite{touvron2023llama}, researchers have increasingly focused on extending their capabilities to the multimodal domain, giving rise to MLLMs.  
Related works in this field have primarily advanced along two complementary axes: architectural design and training strategies. On the architectural side, early frameworks such as BLIP-2 \cite{li2023blip}, Flamingo \cite{alayrac2022flamingo}, and LLaVA \cite{liu2023visual} adopted a connector-based paradigm, coupling a pre-trained visual encoder with a frozen LLM via lightweight projection or query modules to align visual and textual embeddings.
Subsequent works moved toward end-to-end co-training to strengthen multimodal grounding, as exemplified by VILA \cite{lin2024vila} and Qwen-VL \cite{wang2024qwen2}. More recently, there has been a growing interest in developing MLLMs that can jointly perform visual understanding and generation within a unified model \cite{team2024chameleon,chen2025janus}.
For example, researchers explored encoding visual inputs into discrete tokens and generating images in a fully autoregressive manner in MLLMs \cite{wang2024emu3,qu2025tokenflow}.
On the training side, current MLLMs commonly followed a three-stage pipeline: large-scale multimodal pretraining on image-text corpora, supervised instruction tuning (SFT) with curated multimodal tasks (\textit{e.g.}, VQA, captioning, grounding), and preference alignment to refine output helpfulness, factuality, and safety \cite{ouyang2022training,wang2024lift,zhang2025r1,wang2025unified}. Methods such as RL have recently been adapted for multimodal contexts to align MLLMs with human preferences \cite{ahn2024tuning,wang2025rovrm,wang2025vl}. This work belongs to the latter, where our MSRL approach can enhance the performance of MRMs, thereby improving preference alignment in MLLMs.

\subsection{Multimodal Reward Modeling}
MRMs, typically initialized from an MLLM and trained on human preference data, are central to RL with human feedback and other alignment strategies, such as rejection sampling \cite{wang2025rovrm,zang2025internlm,shu2025large}.  
A notable recent advance in MRM research was the paradigm shift from discriminative to generative reward modeling, paralleling developments in LLMs.
This transition aimed to leverage MLLMs' inherent generative and reasoning capabilities to build more powerful and generalizable MRMs \cite{mahan2024generative,wang2025rovrm,xiong2025llava,wang2025gram,zhang2025basereward}.
Furthermore, with the emergence of unified multimodal LLMs \cite{team2024chameleon,chen2025janus}, researchers sought to develop unified MRMs capable of evaluating both understanding and generation tasks \cite{wang2025unifiedcot,wang2025unified,unifiedreward-flex}.
These models not only assessed response quality in image understanding (\textit{i.e.}, ranking textual outputs given an image) but also evaluated visual generations (\textit{i.e.}, comparing images produced from text-to-image prompts).
More recently, inspired by the success of RLVR in eliciting stronger reasoning abilities, there was growing interest in training MRMs with RLVR, where preference labels were used to construct verifiable reward signals \cite{wang2025unifiedcot,zhang2025r1}.
However, this training process remains difficult to scale, primarily because labeled multimodal preference data are scarce.

\section{Preliminaries}

\subsection{Multimodal Reward Modeling}
MRMs play a central role in aligning MLLM outputs with human preferences. Typically, an MRM is initialized from an MLLM and parameterized as $r_{\phi}(x, y)$, where $\phi$ denotes model parameters, $x$ is the multimodal input, and $y$ is the model-generated response.
Here, we consider two main categories of multimodal tasks, including visual understanding and visual generation. For visual understanding, the input $x$ consists of a textual query paired with an image or video. For example, given an image and the query ``\textit{What is the attire of the kayaker?}'', the response $y$ is a sequence of tokens produced by the MLLM, such as ``\textit{Yellow}''. For visual generation, the input is a textual description, and the output is the corresponding visual content. For example, given the caption ``\textit{Generate an image of a cat}'', the response $y$ refers to the generated image.
Under this formulation, the goal of an MRM is to assign a scalar reward that reflects the human preference quality of the output $y$ conditioned on the multimodal input $x$, regardless of whether the task involves textual reasoning over visual content or the evaluation of generated images or videos. To date, mainstream multimodal reward modeling can be broadly categorized into two types: \textit{discriminative} and \textit{generative}.

\vspace{-4mm}
\paragraph{Discriminative Multimodal Reward Modeling.}
Discriminative MRMs compute scores directly as scalar outputs from a classification architecture. The concatenated input–response $[x, y]$ is passed via a pre-trained MLLM, and the final-layer hidden representations are used to compute a scalar score. This model can be trained through a Bradley-Terry loss function \cite{bradley1952rank}:
\begin{eqnarray}
\mathcal{L}_{\mathrm{d}} & = & - \mathbb{E}_{\tiny (x,y_{a},y_{b})\sim D_{r}} \nonumber \\ 
& & \big[ \log (\sigma (r_{\phi}(x,y_{a})-r_{\phi}(x,y_{b}))) \big]
\end{eqnarray}
where $D_{r}$ is the training dataset consisting of tuples of input $x$ and response pair $(y_{a},y_{b})$ with the preference $y_{a} \succ y_{b}$. Once trained, the MRM can be used as a scoring function that assigns a numerical reward $r_{\phi}(x, y)$ to a response $y$ given its corresponding input $x$. 

\vspace{-4mm}
\paragraph{Generative Multimodal Reward Modeling.}
Generative MRMs produce reward signals via text generation. 
Specifically, they leverage an MLLM to generate preference-related tokens, given a natural language prompt $p$ and a tuple $(x, y_a, y_b)$ \cite{wang2025gram,wang2025gram-gen1, mahan2024generative}. The prompt describes the task in natural language, and the model predicts a label token $w$ that aligns with the human preference $l$, where $l = \mathrm{A}$ denotes preference for $y_a$, and $l = \mathrm{B}$ indicates preference for $y_b$. The model can be trained by
\begin{eqnarray}
\mathcal{L}_{\mathrm{g}} &= & - \mathbb{E}_{(p,x,y_a,y_b,l) \sim D_r}
\big[\log \pi_{\phi}(w=l|s)\big] \label{eq:gem-reward-modeling}
\end{eqnarray}
where $s$ denotes the string $[p,x,y_a,y_b]$, and $\pi_{\phi}(\cdot)$ denotes the probability of token prediction by the MLLM. 
More recently, to better exploit MLLMs' inherent reasoning capabilities, researchers have incorporated chain-of-thought reasoning into generative MRMs \cite{wang2025unifiedcot,zhang2025r1}. In this formulation, the model output can be denoted as $o$, and consists of two key components, marked by the tags \texttt{<think>} and \texttt{<answer>}. The \texttt{<think>} section corresponds to the explicit reward reasoning process (\textit{e.g.}, analyzing and evaluating each candidate response), while the \texttt{<answer>} section contains the final preference token $w$.

\begin{figure*}[!t]
    \centering
    \includegraphics[width=1.0\linewidth]{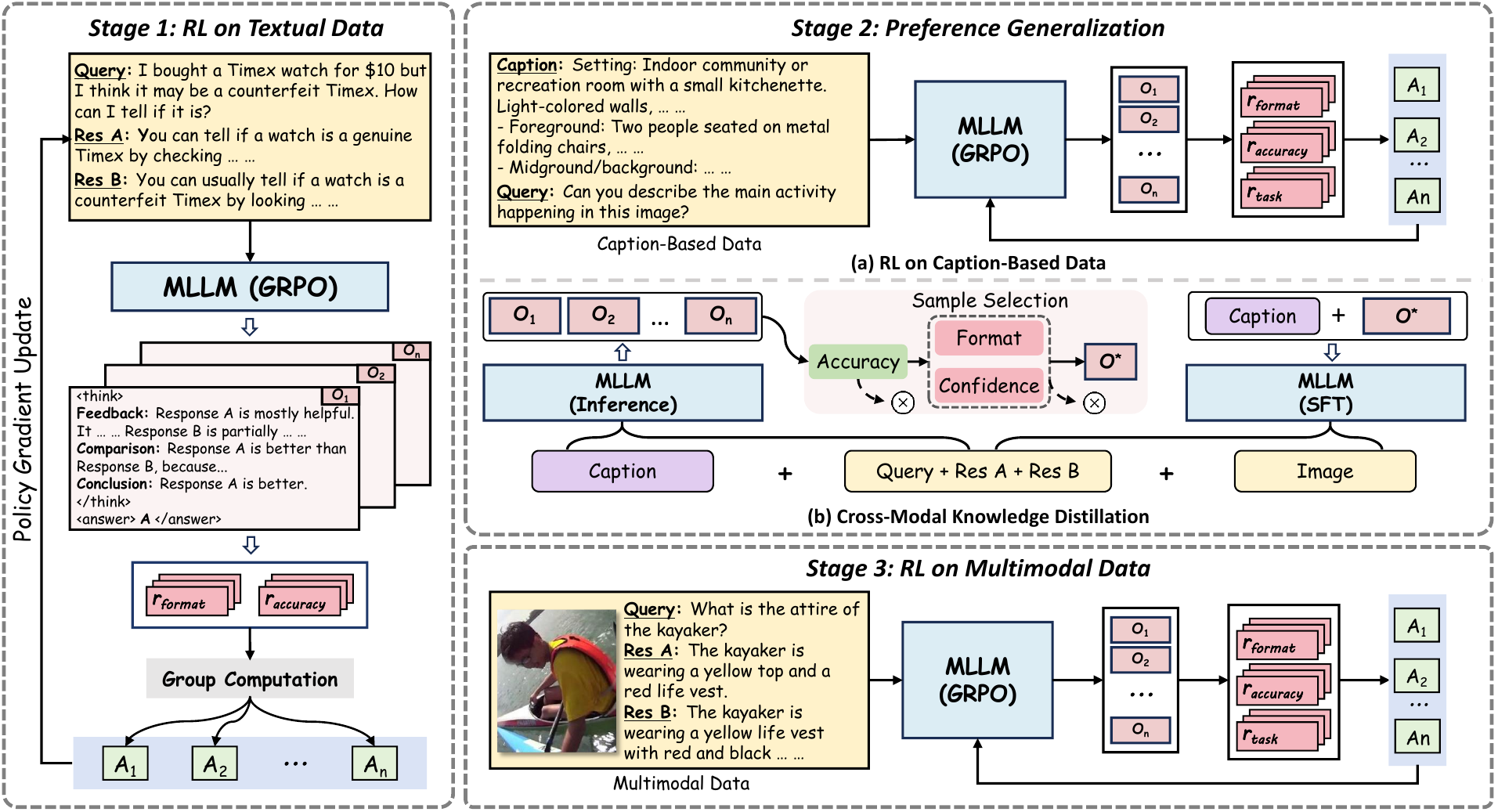}
    \vspace{-4mm}
    \caption{
    An overview of the MSRL approach. We begin by applying RL to large-scale textual preference data (400k examples) to capture rich textual preferences. We then train an RL agent on caption-based data to generalize these preferences to multimodal tasks. During this stage, we also fine-tune the MRMs with CMKD to enhance the generalization. Subsequently, we perform RL with a limited amount of multimodal data to enable adaptation. Note that although the illustration uses image understanding as an example, MSRL is a general approach and can be applied to develop MRMs for arbitrary multimodal tasks.
    }
    \vspace{-2mm}
    \label{fig:main_figure}
\end{figure*}

\subsection{Training MRMs with Reinforcement Learning}
We can employ RL to enhance MRMs' reward reasoning capabilities. RLVR provides a practical mechanism for this purpose. To apply RLVR, we typically define two verifiable rewards: a format reward $r_{\mathrm{format}}(s, o)$ and an accuracy reward $r_{\mathrm{accuracy}}(s, o)$. The format reward $r_{\mathrm{format}}$ is set to 1 if the model output conforms to the required output structure, and 0 otherwise. Similarly, the accuracy reward $r_{\mathrm{accuracy}}$ is set to 1 when the predicted preference token matches the ground-truth preference, and 0 otherwise. Based on the designed verifiable rewards, we optimize the MRM using the following objective:
\begin{eqnarray}
    \mathcal{L}_{\mathrm{RLVR}} &=& - \mathbb{E}_{(p,x,y_a,y_b,l) \sim D_r, o\sim \pi_{\theta}} [r_{v}(s, o)] \nonumber \\ && - \beta \mathbb{D}_{\mathrm{KL}}(\pi_{\theta}||\pi_{\theta_{\mathrm{old}}})
    \label{eq:rlvr}
\end{eqnarray}
where $\mathbb{D}_{\mathrm{KL}}(\cdot)$ denotes the KL divergence, and $\pi_{\theta_{\mathrm{old}}}$ is the model checkpoint prior to the RLVR update. The verifiable reward $r_{v}(\cdot)$ is defined as the sum of the format reward and the accuracy reward: 
\begin{eqnarray}
r_{v}(s,o)&=&r_{\mathrm{format}}(s,o)+r_{\mathrm{accuracy}}(s,o)
\end{eqnarray}
In practice, we adopt group relative policy optimization (GRPO) \cite{shao2024deepseekmath} to optimize the objective, and the implementation details are provided in Appendix~A. However, despite RLVR's effectiveness in improving MRMs, it introduces a non-trivial limitation: obtaining verifiable rewards requires costly human-annotated multimodal preference data, making it difficult to scale multimodal reward modeling.

\section{Multi-Stage Reinforcement Learning}
In this section, we introduce a \textbf{\underline{M}}ulti-\textbf{\underline{S}}tage \textbf{\underline{R}}einforcement \textbf{\underline{L}}earning (MSRL) approach, which is designed to scale up multimodal reward modeling through multiple stages of RL. An overview of MSRL is shown in Figure~\ref{fig:main_figure}. As illustrated, MSRL first learns a generalizable reward reasoning ability from large-scale textual preference data and then progressively transfers these preferences to multimodal tasks via caption-based and fully multimodal RL stages.

\subsection{Reinforcement Learning on Textual Data}
To overcome the bottleneck introduced by the limited availability of multimodal preference data, MSRL begins by leveraging abundant textual preference data to strengthen reward reasoning before any multimodal learning occurs.
Specifically, we first conduct supervised fine-tuning (SFT) to teach the MRM to generate well-structured chain-of-thought outputs and follow the required output format (\textit{e.g.}, correct placement of the \texttt{<think>} and \texttt{<answer>} tags). In this process, we adopt publicly available datasets that provide complete rationales\footnote{https://huggingface.co/datasets/nvidia/HelpSteer3}. Following SFT, we apply RLVR using Eq.~(\ref{eq:rlvr}) on textual preference data to strengthen the MRM’s reward reasoning capability further. In this process, we use the publicly available dataset released in GRAM-R$^2$, which contains approximately 1M high-quality and labeled textual preference examples\footnote{https://huggingface.co/datasets/wangclnlp/GRAM-RR-TrainingData}. The objective of this stage is to acquire a robust and broadly generalizable textual reward policy that serves as a strong foundation for subsequent cross-modal transfer.
Note that during this text-only training stage, we freeze all parameters of the vision encoder and the projector, since all supervision signals originate exclusively from textual data.

\begin{table*}[!t]
    \centering
    \resizebox{\linewidth}{!}{
    \begin{tabular}{lccc}
\toprule[1.1pt]
Task & Input Content  & Description \\ \midrule
Image Understanding & \fontsize{9}{9} (\texttt{<Query>}, \texttt{<Image>}, \texttt{<Res A>}, \texttt{<Res B>}) & \parbox{8.2cm}{Given an image and a textual query, the model evaluates which of the two responses better answers the query.} \\ \midrule
Image Generation   & \fontsize{9}{9} (\texttt{<Caption>},\texttt{<Image A>},\texttt{<Image B>}) & \parbox{8.2cm}{Given an image caption, the model decides which of the two generated images better reflects the described content.}  \\ \midrule
Video Understanding & \fontsize{9}{9}(\texttt{<Query>},\texttt{<Video>},\texttt{<Res A>},\texttt{<Res B>}) & \parbox{8.2cm}{Given a video and a textual query, the model decides which of the two responses better answers the query.}  \\ \midrule
Video Generation    & \fontsize{9}{9} (\texttt{<Caption>},\texttt{<Video A>},\texttt{<Video B>}) & \parbox{8.2cm}{Given a video caption, the model evaluates which of the two generated videos better reflects the described content.}  \\
\bottomrule[1.1pt]
\end{tabular}}
    \caption{
    Description of different multimodal tasks for generative MRMs. Templates for each task are provided in Appendix A.
    }
    \vspace{-2mm}
    \label{tab:multimodal_task}
\end{table*}

\subsection{Preference Generalization}
Ideally, after large-scale RL on textual preference data, the MRM should generalize the learned preferences to multimodal tasks. In practice, however, this generalization is hindered by two barriers: a \textit{task gap} (differences in task structure) and a \textit{modality gap} (differences between textual and multimodal inputs).
Addressing these gaps, MSRL introduces two key components, described below.

\vspace{-4mm}
\paragraph{Reinforcement Learning on Caption-Based Data.} 
\label{sec:caption-based-training}
Inspired by \citet{liu2023visual} and \citet{wang2025rovrm}, who show that caption-based data can effectively transfer textual knowledge to visual tasks in vision–language models, we address the task gap by performing RL on caption-based preference data. 
Here, we construct this caption-based dataset from the original multimodal preference data by replacing each image or video with its corresponding caption. In doing so, the supervision remains purely textual while preserving the semantic grounding of the underlying multimodal scenario, making it an effective intermediate step for transferring preferences. During this stage, we continue to optimize the MRM using the RLVR objective in Eq.~(\ref{eq:rlvr}), with one modification: we introduce a task recognition reward $r_{\mathrm{task}}$, which encourages the MRM to recognize the task type before performing reward reasoning. Specifically, the model is prompted to output a task tag (\textit{e.g.}, \texttt{<type>Image Understanding</type>}) before generating its rationale.
If the predicted task type matches the ground-truth task, the model receives a reward of 0.2; otherwise, the reward is 0.
This mechanism helps the MRM differentiate among multimodal tasks (see Table~\ref{tab:multimodal_task}) and adjust its reasoning accordingly.
Additionally, to mitigate catastrophic forgetting of the textual reward reasoning capability acquired in Stage~1, we employ an experience replay strategy that mixes a subset of high-quality textual preference samples from the previous stage into each training batch. We denote the MRM trained through this process as $\pi_{\theta_{\mathrm{text}}}$.
             
\vspace{-4mm}
\paragraph{Cross-Modal Knowledge Distillation.}
To address the modality gap, we introduce a cross-modal knowledge distillation (CMKD) approach that aligns textual and multimodal reward reasoning.
Specifically, given a preference sample $s$ and its caption $c$, we first query the caption-trained MRM $\pi_{\theta_{\mathrm{text}}}$ to generate $n$ sampled rationales:
\begin{eqnarray}
\{o_1, o_2, \ldots, o_n\} \sim \pi_{\theta_{\mathrm{text}}}(\cdot \mid s, c).
\end{eqnarray}
We then distill these rationales into a single high-quality supervisory signal $o^{*}$ through a three-step selection process: 1) We compute a pseudo-label $\hat{w} = \mathrm{mode}({w_j}_{j=1}^n)$ where $\mathrm{mode}(\cdot)$ is the voting function, and discard all rationales whose predicted label differs from $\hat{w}$; 2) Among the remaining rationales, those that do not conform to the required output format are filtered out; 3) From the filtered set, we choose the rationale with the highest confidence.
The selected rationale $o^{*}$ is used as the distilled ``teacher'' signal. Finally, we conduct SFT on the pairs $[c, o^{*}]$, encouraging the MRM to reproduce the distilled reasoning process and preference prediction even when conditioned solely on the visual input, \textit{i.e.}, $- \log \pi_\theta([c, o^{*}] \mid s)$. 
Note that we also incorporate the caption $c$ during prediction. We aim to allow the model to inherit the caption information, thereby ensuring consistency in behavior across both textual and visual inputs. Building this setting, in the subsequent RL stage, we change the rationale format. Specifically, the model first generates the visual caption (tagged as \texttt{<caption>}) before performing reward reasoning.

\subsection{Reinforcement Learning on Multimodal Data}
The goal of RL in this stage is to fully adapt the model to handle multimodal tasks. We can do this by Eq.~(\ref{eq:gem-reward-modeling}). In this stage, we also use the task recognition reward described in Section~\ref{sec:caption-based-training}. 
Note that as the preceding stages have already endowed the model with strong reward reasoning capabilities and partial multimodal generalization, this final multimodal RL stage becomes easier and requires much less multimodal data. 
In this work, we consider that since textual preference data are abundant, large-scale RL can be performed in earlier stages. This introduces a readily accessible axis for scaling generative multimodal reward modeling: by scaling RL on textual data, we can continually strengthen the model’s reward reasoning capability, thereby improving multimodal reward modeling without requiring additional multimodal data.

\begin{table*}[!t]
    \centering
    \resizebox{0.98\linewidth}{!}{
\begin{tabular}{lrccccccccc}
\toprule[1.1pt]
\multirow{2}{*}{Method} & \multirow{2}{*}{Params.} & \multicolumn{4}{c}{VL-RewardBench}   & \multicolumn{5}{c}{Multimodal RewardBench}    \\ \cmidrule(l){3-6} \cmidrule(l){7-11}
&   & Gen.   & Hall. & Reason.  & Avg.  & Know.  & Safe.     & VQA  & Other & Avg.     \\  \midrule
\rowcolor{blue!10}
\multicolumn{11}{c}{\textit{MLLM-as-a-Judge}}  \\ \midrule
GPT-4o$^\ddagger$     & - &49.1  &67.6 &70.5 &62.4 & 72.0 & 74.8 & 87.2 & 65.4 & 74.9   \\
GPT-5-mini & - &60.2  &76.5 & 76.0 &70.9 & 76.8 & 50.4 & 51.2 & 70.6 & 62.3  \\
Gemini-1.5-Pro$^\ddagger$  &-  &50.8 & 72.5 &64.2 &62.5 & 66.3 & 94.5 & 87.2 & 63.9 & 63.0   \\
Gemini-3-Pro  &- &54.6  &78.9 &86.7 & 73.4 & 94.2 & 94.7 & 93.0 & 82.4 & 91.1  \\
Claude-3.7-Sonnet$^\ddagger$  &- & 68.1 & 70.7 & 60.8 & 66.5 & 78.1 & 72.0 & 86.8 & 66.7  & 75.9  \\
Qwen3-VL-8B-Instruct& 8B &39.0  &52.0  &64.2  &51.7  & 69.7 & 82.3 & 81.3 & 60.5 & 73.5 \\
InternVL3.5-8B     & 8B & 53.3 & 68.0 & 36.8 & 52.7 & 66.5 & 90.6 & 83.4 & 62.6 & 75.8 \\
InternVL3.5-14B     & 14B & 56.8 & 65.4 & 41.4 & 54.5 & 73.5 & 79.9 & 72.4 & 67.8 & 73.4 \\
\midrule
\rowcolor{green!10}
\multicolumn{11}{c}{\textit{Open-Source Multimodal Reward Models}}    \\ \midrule
MM-RLHF-Reward$^\ddagger$ & 7B & 45.0 & 50.5 & 57.6 & 51.0 & 54.3 & 92.9 & 76.8 & 64.7 & 72.2 \\
R1-Reward$^\ddagger$ & 7B & 63.8 & 85.7 & 64.8 & 71.4 & 74.9 & 99.6 & 86.5 & 58.8 & 80.0  \\
UnifiedReward$^\dagger$   & 7B & 76.5& 70.5 & 65.4 & 70.8 & 58.8 & 64.7 &  68.5 & 56.4 & 62.1  \\
UnifiedReward-think (w/ CoT)$^\dagger$ & 7B & 77.9 & 72.7 & 66.0 & 72.2 & 63.4 & 81.5 & 78.5 & 66.4 & 72.5 \\
LLaVA-Critic$^\ddagger$  & 7B & 54.6 & 38.3 & 59.1 & 50.7 & 56.0 & 70.7 & 69.4 & 59.7 & 64.0  \\ \midrule
\rowcolor{yellow!40}
\multicolumn{11}{c}{\textit{Training on the Same Multimodal Preference Data (InternVL3.5-4B)}}   \\ \midrule
Discriminative MRM     & 4B  & 54.4 & 66.8 & 62.4 & 61.2 & 64.2 & 80.3 & \textbf{85.8} & 54.6 & 71.2 \\  
Generative MRM         & 4B  & 56.6 & 65.4 & 59.4 & 60.5 & 65.7 & 77.2 & 82.8 & 51.5 & 69.3 \\
MSRL (Ours)            & 4B  & 62.4 & 80.5 &  66.3 & 69.7 & 69.8 & 85.4 & 84.3 & 56.2 & 73.9 \\
\ \ \ \ +\textit{voting@16} & 4B & \textbf{63.5} & \textbf{82.2} & \textbf{67.6} & \textbf{71.1} & \textbf{70.7} & \textbf{87.2} & 84.7 & \textbf{56.8} & \textbf{74.9}   \\
\midrule
\rowcolor{yellow!40}
\multicolumn{11}{c}{\textit{Training on the Same Multimodal Preference Data (InternVL3.5-8B)}}   \\ \midrule
Discriminative MRM   & 8B  & 54.8 & 75.6 & 62.4 & 64.3 & 68.4 & 88.6 & 84.2 & 58.5 & 74.9 \\
Generative MRM       & 8B  & 57.2 & 74.5 & 68.2 & 66.6 & 67.6 & 84.3 & 87.3 & \textbf{65.7} & 76.2 \\
MSRL (Ours)          & 8B  & 75.4 & 78.2 & 74.2 & 75.9 & 72.8 & 96.4 & 88.6 & 64.2 & 80.5 \\
\ \ \ \ +\textit{voting@16} & 8B & \textbf{76.5} & \textbf{79.3} & \textbf{76.8} & \textbf{77.5} & \textbf{74.3} & \textbf{96.7} & \textbf{89.2} & 65.1 & \textbf{81.3} \\
\bottomrule[1.1pt]
\end{tabular}}
    \vspace{-1mm}
    \caption{
    Accuracies (\%) on VL-RewardBench and Multimodal RewardBench. The best result in each group is shown in bold. Results marked with $\dagger$ are taken from \citet{wang2025unifiedcot} on VL-RewardBench, while those marked with $\ddagger$ for both VL-RewardBench and Multimodal RewardBench are from \citet{zhang2025r1}. All remaining baseline results are obtained by their publicly available models. For Multimodal RewardBench, the ``Other'' column reports the average accuracy across the general and reasoning subsets.
    }
    \vspace{-2mm}
    \label{tab:main-res-for-image-understanding}
\end{table*}

\section{Experiments}
We evaluated MSRL on unified MRMs for both visual understanding and generation, an especially challenging setting due to the limited availability of high-quality unified multimodal preference data.

\subsection{Experimental Setups}
\paragraph{Model Backbones.}
For our experiments, we used the InternVL3.5 model \cite{wang2025internvl3} as the backbone MLLM for training MRMs. We evaluated models of different scales, including 1B, 4B, 8B, and 14B parameters, to assess the effectiveness and scalability of MSRL across varying model sizes.

\begin{table*}[!t]
    \centering
    \resizebox{0.86\linewidth}{!}{
    \begin{tabular}{lrccccc}
\toprule[1.1pt]
\multirow{2}{*}{Method} & \multirow{2}{*}{Params.} & Image Gen. & Video Under. & \multicolumn{2}{c}{Video Gen.} &  \multirow{2}{*}{Average} \\ \cmidrule(l){3-3} \cmidrule(l){4-4} \cmidrule(l){5-6}
& & GenAIBench &  ShareGPT & GenAIBench  &  VideoGen          \\  \midrule
\rowcolor{blue!10}
\multicolumn{7}{c}{\textit{MLLM-as-a-Judge}}  \\ \midrule
GPT-4o              &- &59.6  &54.7  &63.5    &62.1 &60.0           \\
InternVL3.5-8B   &8B & 56.7  & 58.8  &60.1    &54.5  & 57.5 \\
InternVL3.5-14B  &14B &60.4  &64.3  &66.5    &60.1  & 62.8    \\  \midrule
\rowcolor{green!10}
\multicolumn{7}{c}{\textit{Open-Source Multimodal Reward Models}}  \\ \midrule
VisionReward$^\dagger$ & 19B & 66.4 & - & 73.1 & 68.2 & -  \\
UnifiedReward$^\dagger$  &7B & 70.9 & 72.6 & 77.2 & 79.3 & 75.0       \\
UnifiedReward-think (w/ CoT)$^\dagger$ &7B &72.5  &81.0   &82.3  &80.5 & 79.1       \\ \midrule
\rowcolor{yellow!40}
\multicolumn{7}{c}{\textit{Training on the Same Multimodal Preference Data (InternVL3.5-4B)}} \\ \midrule
Discriminative MRM &4B & 69.0  & 72.2  & 76.4  & 73.2 & 72.7 \\
Generative MRM     &4B & 62.3  & 70.4  & 75.9  & 71.6 & 70.1 \\
MSRL (Ours)        &4B & 70.4  & 76.5  & 80.6  & 78.5 & 76.5 \\
\ \ \ \ +\textit{voting@16} & 4B & \textbf{71.8} & \textbf{78.4} & \textbf{81.8} & \textbf{79.2} & \textbf{77.8}          \\ \midrule
\rowcolor{yellow!40}
\multicolumn{7}{c}{\textit{Training on the Same Multimodal Preference Data (InternVL3.5-8B)}}   \\ \midrule
Discriminative MRM   &8B &72.3  &74.5  &70.3    &76.4 &73.4      \\
Generative MRM       &8B &70.2  &80.6  &68.3    &78.2 &74.3            \\ 
MSRL (Ours)          &8B &75.7  &85.5  &81.4    &80.1 & 80.7          \\ 
\ \ \ \ +\textit{voting@16} & 8B & \textbf{75.9} & \textbf{87.3} & \textbf{82.5} & \textbf{82.0} & \textbf{81.9}         \\ 
\bottomrule[1.1pt]
\end{tabular}}
    \vspace{-1mm}
    \caption{
    Accuracies (\%) on image generation, video understanding, and video generation tasks. Results marked with $\dagger$ are taken from \citet{wang2025unifiedcot}. All remaining baseline results are obtained by their publicly available models.
    }
    \vspace{-2mm}
    \label{tab:main-res-for-other-res}
\end{table*}

\vspace{-4mm}
\paragraph{Training Datasets.}
In Stage 1, we first fine-tuned the MRMs using 40k rationale-based preference examples from HelpSteer3 \cite{wang2025helpsteer3}. We then performed RLVR on 400k preference examples randomly sampled from GRAM-R$^2$ \cite{wang2025gram}.
Note that although GRAM-R$^2$ includes synthetic rationales, we did not use them and instead rely solely on the preference label tokens. This design choice was motivated by our empirical observation that large-scale SFT provides limited generalization benefits for MRMs compared with RL. This trend is consistent with prior findings showing that SFT offers weaker generalization than RL \cite{kirk2023understanding}. In Stage~2, we constructed 20k caption-based preference examples by randomly sampling from multiple sources spanning image understanding, image generation, video understanding, and video generation. For each image or video, we used \texttt{GPT-5} to generate textual captions that serve as the textual grounding for this stage. In Stage~3, we again sampled 20k multimodal examples from the same data sources used in Stage~2, but retained the original visual inputs rather than their captions to perform multimodal RL. A dataset list and the corresponding data distributions are shown in Appendix~A.

\vspace{-4mm}
\paragraph{Settings.}
For all GRPO training runs in MSRL, we used the \texttt{ms-swift} framework. We adopted a sampling size of 8, a learning rate of 1e-6, and a batch size of 128, while keeping all other hyperparameters at their default values. In Stage~2, we additionally employed experience replay during RL training, mixing new and previously seen data at a 5:1 ratio. Further details are provided in Appendix~A.

\vspace{-4mm}
\paragraph{Baselines.}
Our primary baseline was \textit{UnifiedReward} \cite{wang2025unifiedcot}, which is trained using RLVR solely on multimodal preference data. We also compared against several \textit{open-source MRMs}, such as R1-Reward \cite{zhang2025r1} and LLaVA-Critic \cite{xiong2025llava}. Additionally, we included an \textit{MLLM-as-a-Judge} baseline by prompting strong MLLMs to predict preference labels. To control for data and model differences, we further trained both a vanilla discriminative MRM and a generative MRM using the same backbone and the same multimodal preference data, enabling a fair comparison with our approach (denoted as \textit{Discriminative MRM} and \textit{Generative MRM}).

\vspace{-4mm}
\paragraph{Evaluation.}
We followed the evaluation protocol of \citet{wang2025unified} to assess our MRMs.
For image understanding, we evaluated on the VL-RewardBench \cite{li2025vl} and the Multimodal RewardBench \cite{yasunaga2025multimodal}.
For image and video generation, we used GenAI-Bench \cite{jiang2024genai}, and additionally evaluated video generation performance on VideoGen-RewardBench \cite{liu2025improving}.
For video understanding, we used ShareGPTVideo \cite{zhang2025direct}.

\subsection{Results}
\paragraph{Image Understanding and Generation Tasks.}
Table~\ref{tab:main-res-for-image-understanding} reports the performance of MSRL and various baselines on VL-RewardBench and Multimodal RewardBench. A key finding from the results is the consistent and substantial performance improvement brought by incorporating RL on large-scale textual preference data.
We also find that across both backbone settings, MSRL significantly outperforms the discriminative and generative MRM baselines trained on the same multimodal preference data, demonstrating that multimodal reward modeling can be effectively scaled via RL on textual data. 
Furthermore, compared to CoT-based MRMs that rely solely on limited multimodal preference data, MSRL attains stronger performance through a more direct and scalable approach, \textit{i.e.}, simply extending RL to abundant textual preferences without requiring additional costly multimodal annotations.
This highlights both the practicality and scalability of MSRL for training MRMs. Additionally, we observe that applying a simple majority voting strategy (denoted as +\textit{voting@16}) further boosts performance across both backbones, yielding +1.0-1.5\% gains on average (\textit{e.g.}, from 80.5\% to 81.3\% for 8B on the Multimodal RewardBench). Furthermore, we evaluate MSRL on the image generation task using GenAI-Bench, with results summarized in Table~\ref{tab:main-res-for-other-res}.
A similar performance trend is observed, confirming the effectiveness of our MSRL approach across both understanding and generation tasks.

\begin{table}[!t]
    \centering
    \vspace{1mm}
    \resizebox{\linewidth}{!}{

\begin{tabular}{lcccc}
\toprule[1.1pt]
Method & Gen. & Hall. & Reason. & Avg. \\  \midrule
Generative Baseline & 57.2     &74.5       &68.2         &66.6      \\ \midrule
MSRL     &\textbf{75.4}      &\textbf{78.2}       &\textbf{74.2}         &\textbf{75.9}      \\
w/o Stage 1         &62.7      &73.4       &70.2   &  68.8        \\
w/o Stage 2 (Caption Data) &73.2    &76.3       &73.5         &74.3          \\
w/o Stage 2 (CMKD)       &72.4      &75.8       &72.0         &  73.4        \\
w/o Stage 3        &70.8      &74.6       &72.4         &  72.6        \\
\bottomrule[1.1pt]
\end{tabular}}
    \vspace{-2mm}
    \caption{
    Ablation results on VL-RewardBench. It is worth noting that when Stage~1 is removed, the SFT-based cold-start process included in Stage~1 is also removed.
    }
    \vspace{-2mm}
    \label{tab:ablation_study}
\end{table}

\begin{figure*}[!t]
    \centering
    \includegraphics[width=\linewidth]{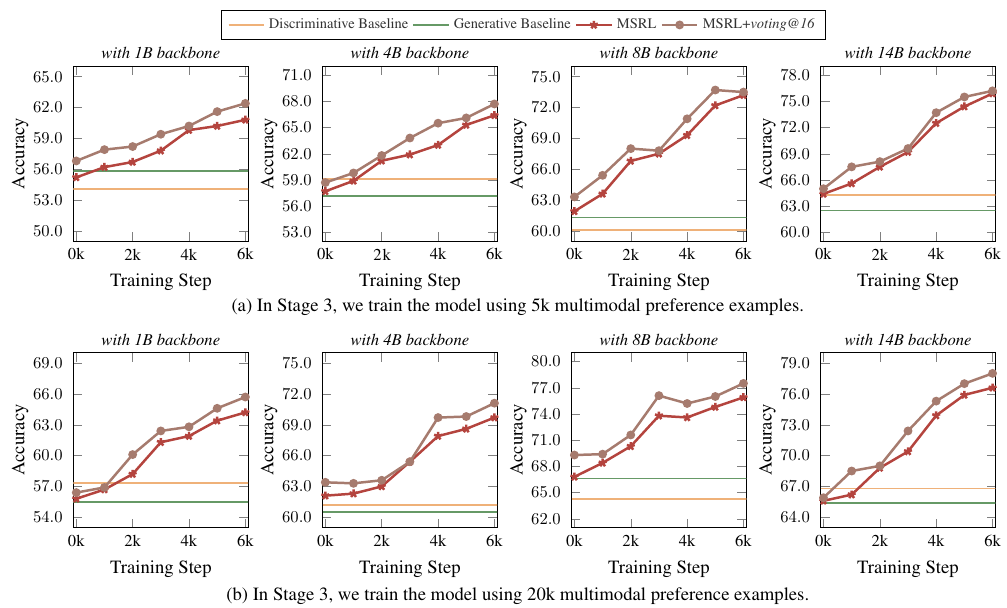}
    \vspace{-7mm}
    \caption{
    Performance scaling with different amounts of textual preference data on the VL-RewardBench.
    }
    \vspace{-2mm}
    \label{fig:scaling_res_with_diff_model_sizes}
\end{figure*}

\vspace{-2mm}
\paragraph{Video Understanding and Generation Tasks.}
The inherent richness of temporal–visual data, coupled with the noise and scarcity of video preference data, makes video understanding and generation substantially more challenging for MRMs than image-based tasks.
As shown in Table \ref{tab:main-res-for-other-res},  MSRL effectively overcomes these challenges, substantially improving the performance of MRMs on both video understanding and video generation tasks.
For instance, with the InternVL3.5-8B backbone, MSRL achieves 85.5\% on ShareGPT (video understanding) and 81.4\%  on GenAI-Bench (video generation), outperforming the vanilla generative MRM by +4.9\% and +13.1\%, respectively. These consistent gains highlight that MSRL not only scales multimodal reward modeling efficiently but also generalizes robustly across multimodal tasks. 

\subsection{Ablation Study}
We conduct ablation experiments to assess the contribution of each stage in MSRL using the InternVL3.5-8B backbone on VL-RewardBench, with results summarized in Table~\ref{tab:ablation_study}.
The findings reveal that incorporating textual reinforcement learning in Stage~1 is essential, yielding a substantial performance gain of +6.9\% and demonstrating the effectiveness of leveraging abundant textual preferences for scaling MRMs. Moreover, we find that the preference transfer stage (Stage~2) is indispensable and its removal causes a clear performance drop (-1.6\% to -2.6\%), underscoring its importance in bridging textual and multimodal domains. Finally, the multimodal adaptation stage (Stage~3) further refines reward reasoning and enhances overall generalization.
Interestingly, even without Stage~3, MSRL still surpasses the generative baseline by a considerable margin, indicating that most of the model's reward reasoning capability is already acquired in the earlier stages.

\subsection{Scaling Behavior of MSRL Across Model Sizes}
Figure~\ref{fig:scaling_res_with_diff_model_sizes} reveals a clear and robust scaling trend: MSRL delivers consistent performance gains across model sizes, from 1B to 14B. Regardless of model capacity, MSRL systematically outperforms both discriminative and generative baselines, demonstrating the benefits of staged textual-to-multimodal RL. This demonstrates that textual RL serves as an effective and scalable axis for strengthening generative MRMs. Furthermore, the comparison between using 5k and 20k multimodal samples in Stage~3 highlights MSRL’s strong low–data efficiency. Even with only 5k multimodal examples, MSRL can achieve substantial improvements. This behavior suggests an important scaling insight: once large-scale textual reward reasoning is established, multimodal learning signal exhibits diminishing returns, and only a small amount of multimodal data is needed to achieve strong performance. 

\section{Conclusions}
We presented MSRL, a multi-stage reinforcement learning approach that scales generative multimodal reward modeling by leveraging abundant textual preference data. MSRL first builds strong reward reasoning capabilities on large-scale textual preferences and then transfers them to multimodal tasks with minimal multimodal data. Through cross-modal distillation and staged RL, MSRL enables efficient and scalable training of generative MRMs and consistently outperforms strong baselines across both visual understanding and visual generation tasks.

\section*{Acknowledgments}
This work was supported in part by the National Natural Science Foundation of China (Nos. U24A20334 and 62276056), the Yunnan Fundamental Research Projects (No.202401BC070021), the Yunnan Science and Technology Major Project (No. 202502AD080014), the Fundamental Research Funds for the Central Universities (Nos. N25BSS054 and N25BSS094), and the Program of Introducing Talents of Discipline to Universities, Plan 111 (No.B16009). We would like to thank the anonymous reviewers and SPC for their valuable comments, which helped improve this paper. We also thank Kaiwei Wang for helping reproduce several baselines used in this work.

{
    \small
    \bibliographystyle{ieeenat_fullname}
    \bibliography{main}
}


\clearpage
\setcounter{page}{1}

\onecolumn
\centerline{
\LARGE \textbf{MSRL: Scaling Generative Multimodal Reward Modeling}
}
\vspace{1mm}
\centerline{
\LARGE \textbf{via Multi-Stage Reinforcement Learning}
}
\vspace{1mm}
\centerline{
\Large Supplementary Material
}
\vspace{2mm}

\appendix

\section{Details of Experiments}
\label{app:details_of_experiments}

\subsection{Datasets}
We describe the data sources used in our training pipeline below. For image understanding, we used the vision-feedback-mix-binarized dataset \cite{wang2025rovrm} (denoted as \textbf{S1}). For image generation, we incorporated preference annotations from the open-image-preferences-v1\footnote{https://huggingface.co/datasets/data-is-better-together/open-image-preferences-v1-binarized} dataset (denoted as \textbf{S2}) and the OpenAI-4o-human-preference dataset\footnote{https://huggingface.co/datasets/Rapidata/OpenAI-4o\_t2i\_human\_preference} (denoted as \textbf{S3}). For video understanding, we used the ShareGPTVideo-DPO dataset~\cite{zhang2025direct} (denoted as \textbf{S4}). For video generation, we used VideoDPO~\cite{liu2025videodpo} (denoted as \textbf{S5}) and the text-2-video-human-preferences dataset\footnote{https://huggingface.co/datasets/Rapidata/text-2-video-human-preferences} (denoted as \textbf{S6}). The statistics for the datasets used in Stages~2 and 3 are provided in Table~\ref{tab:data_statistic}.

\begin{table*}[h]
    \centering
    \begin{tabular}{lccccccc}
\toprule[1.1pt]
\multirow{2}{*}{Training Stage}& \multirow{2}{*}{Total} & IU & \multicolumn{2}{c}{IG} & VU & \multicolumn{2}{c}{VG} \\  \cmidrule(l){3-3} \cmidrule(l){4-5} \cmidrule(l){6-6} \cmidrule(l){7-8}
& & S1 & S2 & S3 & S4 & S5  & S6        \\ \midrule
Stage~2 (Caption-Based Data) &19,442  & 5,000 & 2,500            &2,580  & 4,323  &2,500   & 2,539 \\ 
Stage~3 (Multimodal Data)  &20,038    &8,000  &2,500   &2,639   &2,343  &2,500 &2,056           \\ 
\bottomrule[1.1pt]
\end{tabular}
    \caption{Dataset statistics used in Stage~2 and Stage~3.}
    \label{tab:data_statistic}
\end{table*}

\subsection{Settings}
\paragraph{Discriminative and Generative Baselines.} We trained the discriminative and generative reward model baselines for one epoch using a learning rate of 1e-5 and a batch size of 128. We also experimented with other hyperparameter settings, but did not observe meaningful improvements. For the discriminative baseline, we used the complete set of labeled multimodal preference data to train for one epoch. The training template follows the structure illustrated in Figures~\ref{fig:template_for_iu}, \ref{fig:template_for_ig}, \ref{fig:template_for_vu}, and \ref{fig:template_for_vg}. Note that we did not incorporate rationales during training, as the labeled data lacks such annotations.

\paragraph{MSRL Training.}
In Stage~1 (Text-Only Training), we first performed SFT using a learning rate of 1e-5, a batch size of 128, and 3 training epochs. After SFT, we applied GRPO-based reinforcement learning with a sampling rate of 8 and a learning rate of 1e-6 to strengthen further the model’s reward reasoning ability on large-scale textual preference data. In Stage~2 (Caption-Based RL), we continued training using the RLVR objective on caption-based preference data. The hyperparameters largely followed those used in Stage~1, with a learning rate of 1e-6 and a sampling number of 8 to maintain stability when adapting from purely textual inputs to caption-grounded multimodal scenarios. We additionally incorporated a task-recognition reward to encourage the model to identify the underlying task type before generating its rationale. To mitigate catastrophic forgetting of the textual reward reasoning acquired in Stage~1, we interleaved a replay buffer of high-quality textual preference samples into each training batch. In Stage~3 (Cross-Modal Knowledge Distillation), we trained the model using CMKD to align textual reward reasoning with genuine multimodal inputs. During this stage, we adopted a smaller learning rate of 2e-6 and a sampling number of 8, reflecting the increased difficulty and noise inherent in multimodal preference signals. The teacher model was the caption-based MRM obtained from Stage~2, while the student model received both the multimodal input and the distilled rationales to promote consistent reward reasoning across modalities. Across all stages, GRPO training used a batch size of 128. We froze the vision encoder and projector during text-only training in Stage~1 and unfroze them in Stages~2 and 3 to enable effective cross-modal adaptation.

\paragraph{Evaluation.} 
For evaluation, we primarily used VL-RewardBench \cite{li2025vl} and Multimodal RewardBench \cite{yasunaga2025multimodal} to assess performance on image understanding tasks. These benchmarks cover a wide range of task types, such as visual reasoning, safety, and VQA, providing a comprehensive and realistic assessment of an MRM's performance in practical application scenarios. Additionally, we employed GenAIBench, ShareGPT, and VideoGen to evaluate our MRM on image generation, video understanding, and video generation tasks. These benchmarks are widely used in multimodal alignment research for both training and evaluation, enabling a thorough and holistic assessment of our MRM across diverse multimodal settings.

\begin{table*}[!t]
    \centering
    \begin{tabular}{cccccc}
\toprule[1.1pt]
\textbf{\begin{tabular}[c]{@{}c@{}}Mixing Ratio \\ (Caption : Text)\end{tabular}} & 1:0  & 1:1  & 2:1  & 4:1  & 5:1  \\  \midrule
Accuracy                                                                          & 74.6 & 73.8 & 74.2 & 75.5 & 75.2 \\   
\bottomrule[1.1pt]
\end{tabular}
    \caption{
    Performance of preference generalization under different caption-to-text mixing ratios. The ratio ``1:0'' indicates that Stage~2 training uses only caption-based data, with no text-only preference data mixed in.
    }
    \label{tab:performance_with_diff_data_rations}
\end{table*}

\section{More Analysis}
\subsection{Performance Under Different Mixing Ratios for Preference Generalization}
In the preference generalization stage, our pipeline combines caption-based RL with CMKD. A key challenge in caption-based RL is catastrophic forgetting, where the model overfits to the caption distribution and loses the reward reasoning capability acquired in Stage~1. To assess the effectiveness of different data mixtures, we evaluate several caption-to-text ratios {1:1, 2:1, 4:1, 5:1}. We further construct a caption-based test set by converting VL-RewardBench images into textual descriptions using GPT-5. As shown in Table~\ref{tab:performance_with_diff_data_rations}, insufficient text-only preference signals (e.g., 1:0) lead to clear degradation, while overly balanced mixtures (e.g., 1:1) underexploit caption distributions. The 4:1 ratio achieves the highest accuracy (75.5), suggesting that a caption-leaning mixture best preserves Stage~1 preference reasoning while improving caption-based generalization. Increasing the caption weight further (5:1) offers diminishing gains, indicating that 4:1 provides the most effective trade-off between stability and preference retention.

\clearpage

\begin{figure*}[!t]
    \centering
    \resizebox{0.88\linewidth}{!}{
    \input{images/templates/get_image_understanding_data}
    }
    \caption{Template used for the image understanding task.}
    \label{fig:template_for_iu}
\end{figure*}

\begin{figure*}[!t]
    \centering
    \resizebox{0.88\linewidth}{!}{
    \input{images/templates/get_image_generation_data}
    }
    \caption{Template used for the image generation task.}
    \label{fig:template_for_ig}
\end{figure*}

\begin{figure*}[!t]
    \centering
    \resizebox{0.88\linewidth}{!}{
    \input{images/templates/get_video_understanding_data}
    }
    \caption{Template used for the video understanding task.}
    \label{fig:template_for_vu}
\end{figure*}

\begin{figure*}[!t]
    \centering
    \resizebox{0.88\linewidth}{!}{
    \input{images/templates/get_video_generation_data}
    }
    \caption{Template used for the video generation task.}
    \label{fig:template_for_vg}
\end{figure*}

\begin{figure*}[!t]
    \centering
    \includegraphics[width=0.90\linewidth]{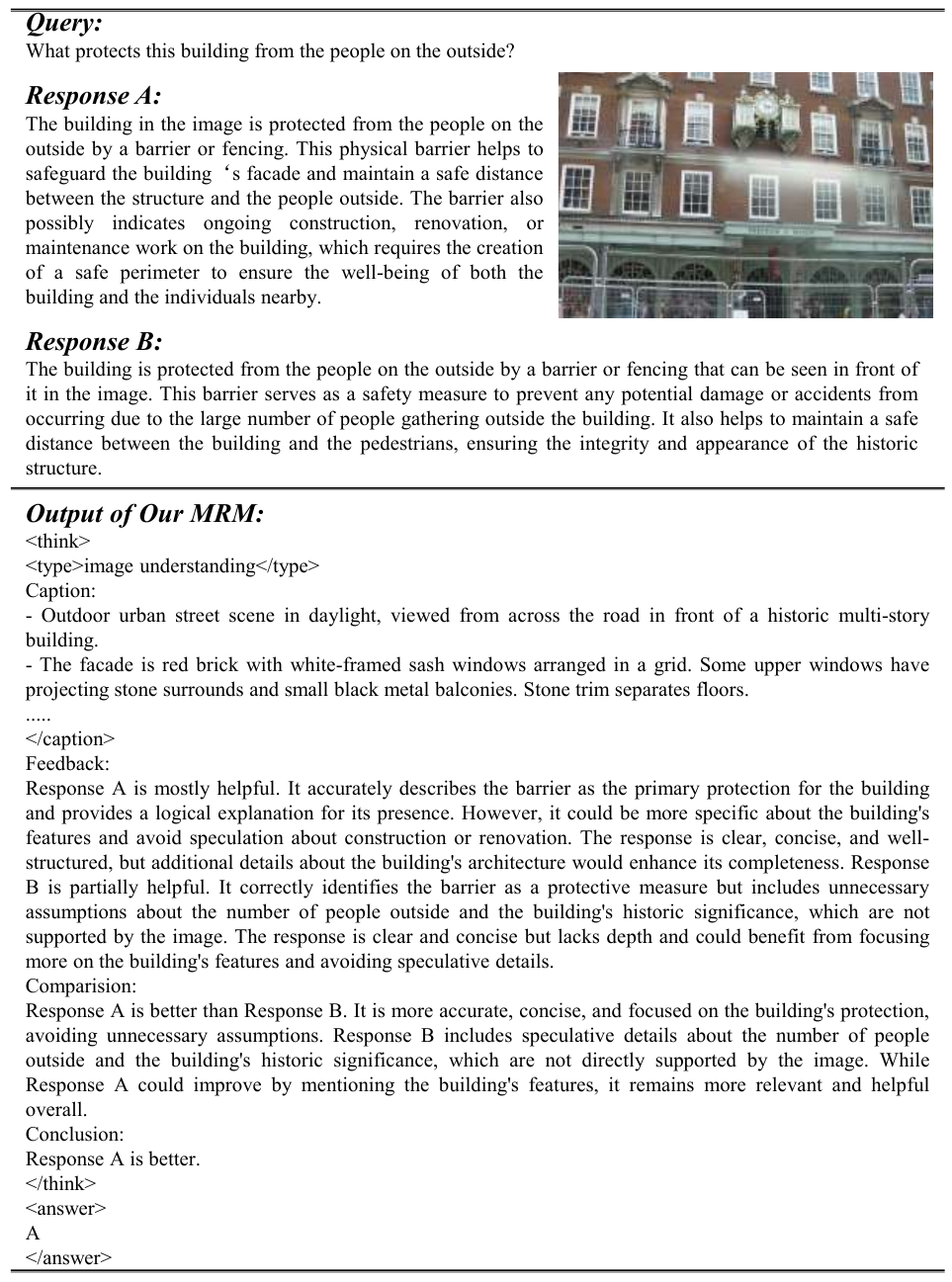}
    \vspace{-4mm}
    \caption{
    A case study illustrating the rationale generated by MSRL (8B backbone) for the image understanding task. Response A is identified as the superior answer because it provides an accurate and concise description grounded in visual evidence, whereas Response B relies on unsupported speculations regarding crowd size and the building's historic significance. This demonstrates MSRL's capability to discern these nuances and predict preferences with high precision effectively.
    }
    \label{fig:case_for_iu}
\end{figure*}

\begin{figure*}[!t]
    \centering
    \includegraphics[width=0.88\linewidth]{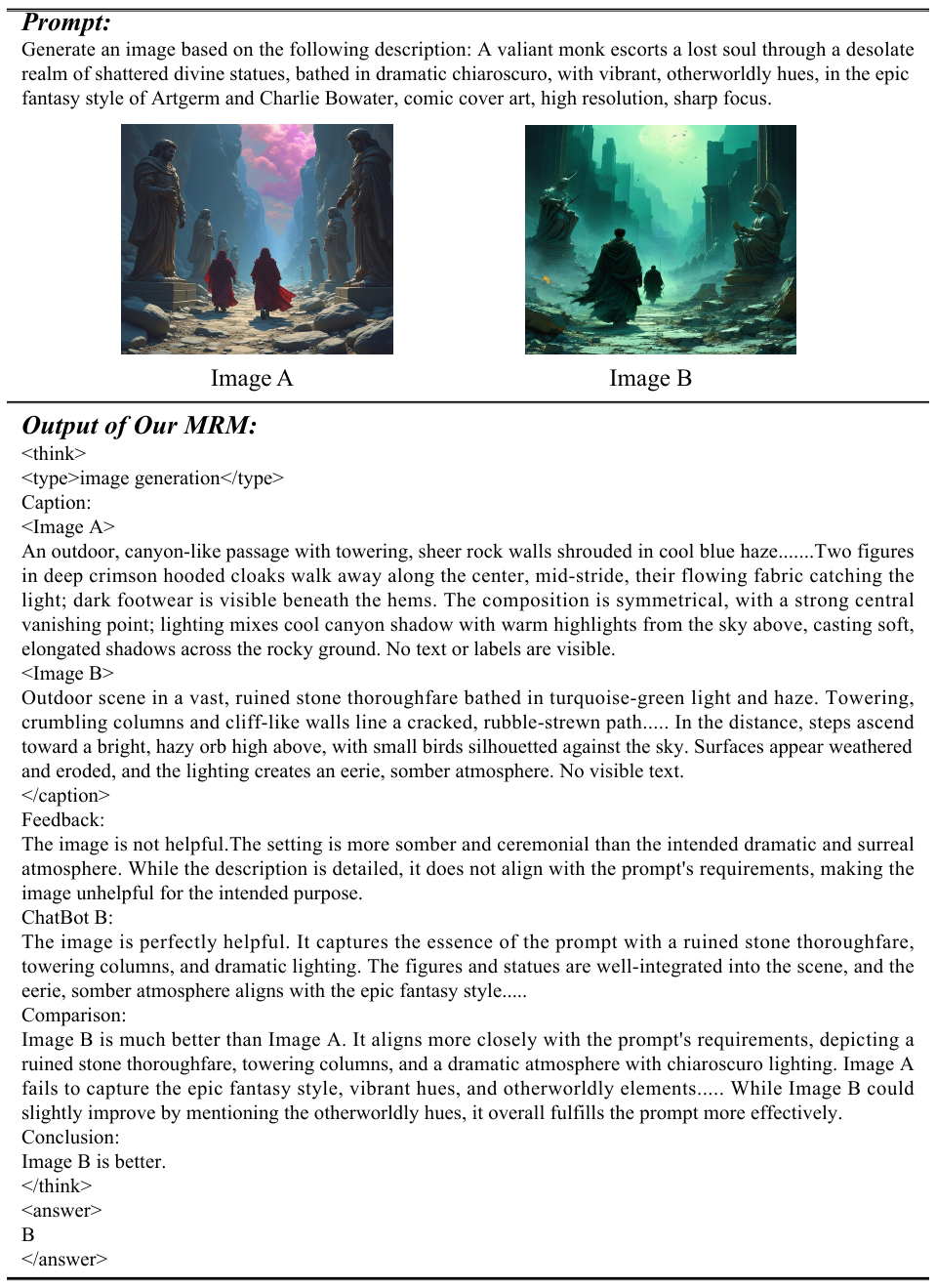}
    \vspace{-4mm}
    \caption{
    A case study illustrating the rationale generated by MSRL (8B backbone) for the image generation preference task. Image B is identified as the superior output because it effectively captures the prompt's request for a ``ruined stone thoroughfare'' and ``epic fantasy style'' with dramatic lighting. In contrast, Image A is critiqued for missing the ``vibrant hues'' and ``otherworldly elements'', presenting a setting that is too somber and ceremonial. This highlights MSRL's ability to evaluate semantic alignment and stylistic fidelity in text-to-image generation precisely.
    }
    \label{fig:case_for_ig}
\end{figure*}

\begin{figure*}[!t]
    \centering
    \includegraphics[width=0.90\linewidth]{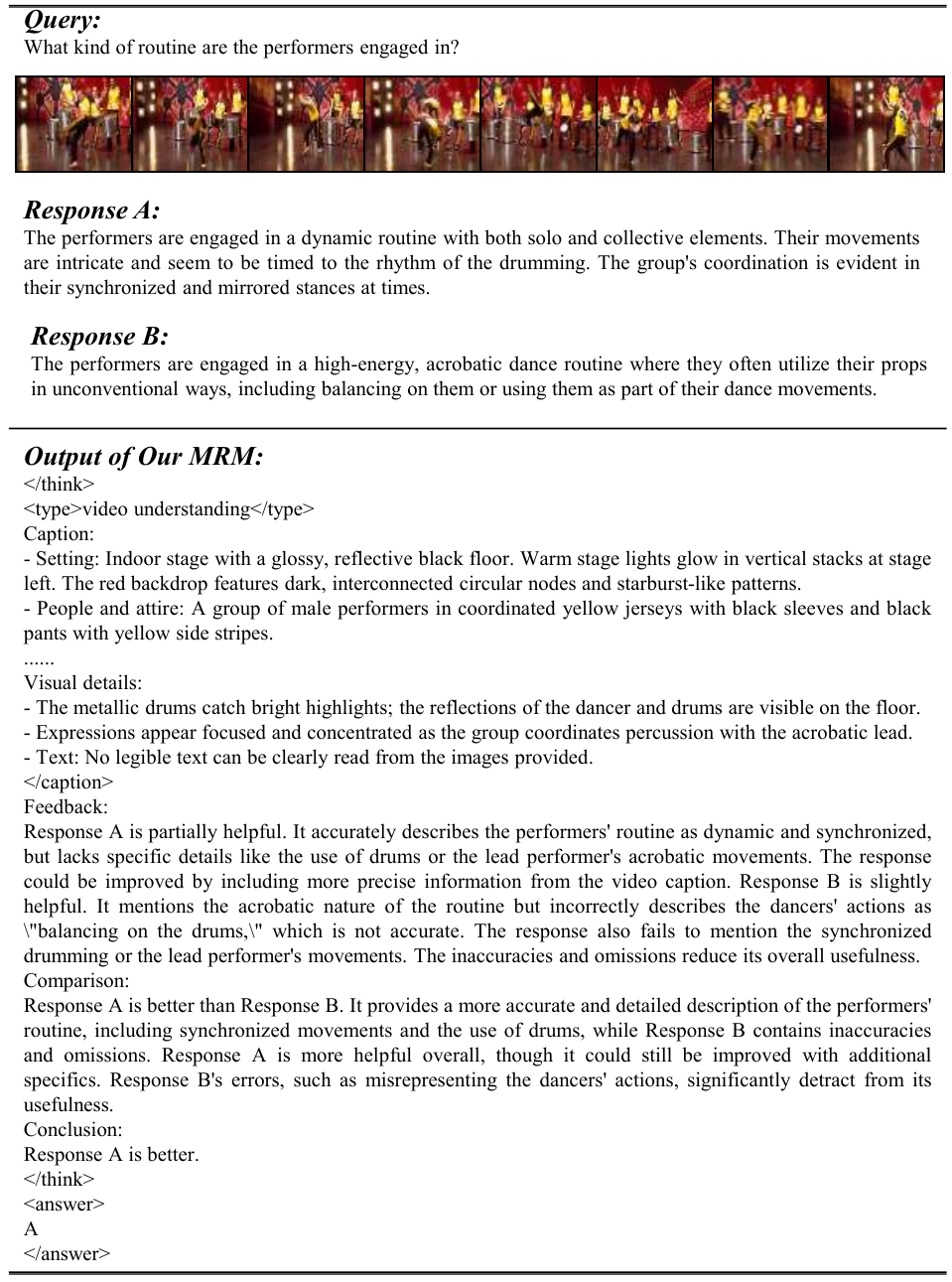}
    \vspace{-4mm}
    \caption{
    A case study illustrating the rationale generated by MSRL (8B backbone) for the video understanding task. Response A is identified as the superior answer because it accurately captures the synchronized movements and rhythmic coordination of the performers. In contrast, Response B is rejected due to visual inaccuracies, specifically the hallucinatory claim that performers are ``balancing on the drums''. This demonstrates MSRL's capability to verify visual details across temporal frames and identify factual inconsistencies.
    }
    \label{fig:case_for_vu}
\end{figure*}

\begin{figure*}[!t]
    \centering
    \includegraphics[width=0.86\linewidth]{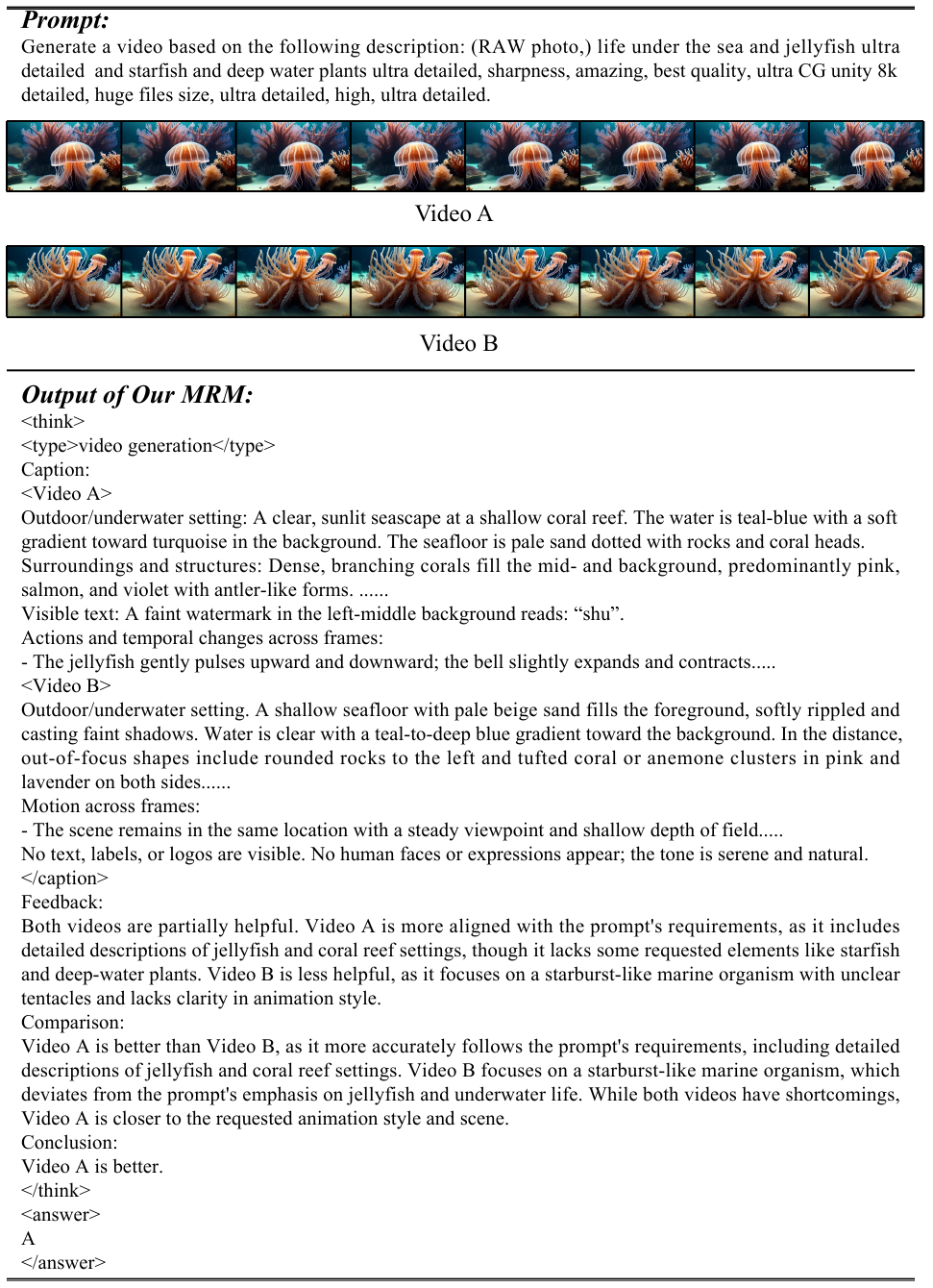}
    \vspace{-4mm}
    \caption{
    A case study illustrating the rationale generated by MSRL (8B backbone) for the video generation preference task. Video A is identified as the superior output because it accurately depicts the primary subject (\textit{i.e.}, a ``jellyfish'') within a detailed coral reef setting, aligning closely with the prompt. In contrast, Video B is critiqued for deviating from the prompt's requirements, focusing instead on an ambiguous ``starburst-like marine organism'' rather than the requested underwater life. This demonstrates MSRL's capability to evaluate subject fidelity and semantic alignment in generated video content.
    }
    \label{fig:case_for_vg}
\end{figure*}

\end{document}